\documentclass[12pt]{article}

\usepackage{sbc-template}

\usepackage{graphicx,url}

\usepackage[utf8]{inputenc} 

\usepackage{tabu}

\title{Uncovering Soccer Teams Passing Strategies Using Implication Rules}

\author{Olumide Leshi\inst{1} }

\address{IDA - Linköping University SE-581 83 Linköping, Sweden
  \email{olule466@student.liu.se}
}

\begin{document} 

\maketitle

\begin{abstract}
Formal Concept Analysis (FCA) has seen application in different knowledge areas, including Social Network Analysis (SNA). In turn, research has also shown the applicability of SNA in assessing team sports. In this project, to uncover frequent passing sequences of a soccer team, an FCA- based approach is introduced. The approach relies on a minimum cover of implications, the Duquenne- Guigues (DG) basis and the notion that a soccer team’s passes describe a social network.
\end{abstract}

\section{Introduction}

A social network is structured to consist of social actors and relations between them, the least of which can be dyadic (binary) \cite{wasserman1994social}. Social Network Analysis (SNA) focuses on studying social networks to reveal axes of influence by each social actor and the generalisation of any such patterns of influence, given the dynamics of the network. It is then easy to think of a soccer team as being a social network, with passes establishing the basis for the relation between players.

Much effort in the analysis of team sports has focused on statistical approaches and the visualisation of such data. However, it is also essential to consider alternative approaches like formal concept analysis. FCA is particularly suited for this purpose since formal concept lattices can express sociograms\protect\footnote{https://en.wikipedia.org/wiki/Sociogram}. Meaning, implication rules can describe a social network.

Concept lattices and implications rules provide views of the concepts encoded by FCA from data. This project focuses on a minimum cover of the set of implications, the Duquenne-Guigues (DG) basis (or the canonical base). The benefit of implications to social network analysis includes the ability to observe social interactions and evolving patterns in the network \cite{aufaure2013advances}. For example,  a team’s sequence of passes can develop into frequent patterns if its style of play is consistent, and this is observable in subsequent matches.

In this project, the approach to identifying a soccer team’s passing strategy relies on the formalism of passing sequences as implications and searching for such implication patterns in subsequent matches. 

The dataset featured is the one published as part of the project, \textit{"A Public Data set of Spatio-Temporal Match Events in Soccer Competitions"}, that has helped to provide an open-access log of spatiotemporal events like passes, shots and fouls \cite{pappalardo2019public} for soccer matches. However, consideration is only for events relating to FC Barcelona, 2017/2018 \textit{La Liga} (Spanish Premier Division) season title winners. Data from the team’s first three league matches of that season helps to demonstrate the approach.

Previous work focusing on the application of formal concept analysis to team sports data is minimal (if at all), and this project contributes to addressing that dearth. Hopefully, the techniques described can spur further research and be useful in analysing passes in other relevant team sports.

Section 3 contains a formal description of implication rules, but presented in Section 4 is an insight into the methodology adopted for this project. Section 5 looks at the results obtained, and a fitting conclusion in is given in Section 6.

\section{Related work} \label{sec:relatedwork}

There exist several research efforts in the application of FCA to social network analysis, including works by \cite{krajvci2014social} \cite{neznanov2015analyzing} \cite{banerjee2014gossip} and \cite{stattner2012social}, in addition to a book by \cite{abraham2009computational}. However, none of these authors treated the matter of analysing social networks in team sports.

The minimal cover of implication basis, Duquenne-Guigues, was used by \cite{cordero2015knowledge} to analyse relationships in the (now defunct) online social network, Orkut. Using real network traffic data, the authors sought a minimal set of user access patterns capable of representing the network structure. The research demonstrated the usefulness of the relations between implication premises and conclusions in representing the (web- site) navigation behaviour of actors in the social network, given available functionalities.

Similar research by \cite{neto2018identification}, \cite{resende2015canonical} and \cite{neto2015using} motivate the application of implication rules in analysing social networks, but in particular the canonical base of implications. The benefit being the compression quality of this set of implications. However, these authors do not consider social networks evident in team sports.

The work by \cite{gyarmati2015automatic} proposed to uncover soccer team passing strategies using Dynamic Time Warping (DTW) and also considers the Spanish topflight division, La Liga. This approach is purely statistics-based; functional data analysis. \cite{van2015automatically} proposed the adoption of data-mining and inductive logic programming to automatically discover frequently observed patterns in a team’s offensive tactics. Of particular interest, the authors note, are offensive moves that might lead to goals. However,  this differs from the approach proposed in this project.

\section{Formal Concept Analysis (FCA)}

A brief introduction to the formalisms related to FCA is provided in this section, courtesy of the work by \cite{ganter2012formal}, \cite{ganter2010two} and \cite{stumme1999hierarchies}.

FCA commences by building a formal context, which encodes the relation between a set of \textit{objects} and their \textit{attributes}, also taken to be a set. From this context are derived \textit{concepts}, that each can be described in terms of \textit{intents} and \textit{extents}. Attributes of a concept form its intent and objects the extent.

From a formal context, there can be two representations of the information encoded: (i.) Concept Lattice (ii.) Implications. In this project, the focus is on generating valid implication rules. The reader is encouraged to refer to \cite{ganter2012formal} for an in-depth take on intents, extents and concept lattices.

\subsection{Formal Context}
A formal context is a triple \textit{(G,M,I)}, where
\textit{G} is a set of objects, and \textit{M} is a set of attributes. The relation \textit{I} is defined over the two sets such that I $\subseteq$ \textit{G X M}. For a pair \textit{(g,m)} $\subseteq$ \textit{I}, it can be read as 'object \textit{g} has attribute \textit{m}'. Shown in Table \ref{table:formalcontexttoysample} is a formal context.

Note that the notion of incidence \textit{I} (i.e \textit{relation}) is what each asterisk in Table \ref{table:formalcontexttoysample} conveys. That is, there is either a pass to the receiving player or not. For example, there is a pass from \textit{Sergio} to \textit{Messi}. The incidence \textit{I} in this case is therefore a Boolean function.

\begin{table}[h!]
\begin{center}
\begin{tabu} { | l | l | l | l | }
 \hline
 \textbf{\textit{Objects}} & \textbf{Messi}  & \textbf{Suarez}  & \textbf{Neymar}\\
 \hline
 Rakitic &  & * & *\\
 \hline
 Sergio & * & * &  \\
 \hline
 Busquet & * & * &  \\
 \hline
 Pique & * &  & * \\
 \hline
 Alba & * &  &  \\
\hline
\end{tabu}\\
\end{center}
\caption{A formal context expressing passing relations between players. The rows indicate \textit{objects}: players initiating the pass; the columns are \textit{attributes} detailing pass-receiving players.}
\label{table:formalcontexttoysample}
\end{table}

\subsection{Conceptual-Scaling of Formal Contexts}
Away from the simple binary formal context shown in Table \ref{table:formalcontexttoysample}, more often than not, data in the real-world can become more complex. In such cases, \textit{I} will likely encode incidence between objects and \textit{attribute-value} pairs.

To attend to this type of structuring, \cite{stumme1999hierarchies} notes that \cite{ganter2012formal} introduced the idea of conceptual-scaling according to which \textit{many-valued} attributes can be dealt with. Hence, so-called \textit{many-valued contexts} can be fleshed-out or 'dissolved' into \textit{single-valued contexts}, like the context shown in Table \ref{table:formalcontexttoysample}.

To motivate the need for conceptual-scaling in this project, consider the following explanation:  Multiple passes to Messi from Sergio can occur in a match, each with different timestamps. In that case, the incidence relation $I_{S,M}$ will have a form like $[Sergio,[\{Messi,4 mins\}, \{Messi, 7 mins\},..,\{Messi, 85 mins\}]]$, with a multi-valued context likely constructed as shown in Table \ref{table:MVformalcontexttoy}.

This type of incidence relation presupposes that the multi-valued context can give rise to a binary single-valued equivalent, but admittedly, one with more attributes. Importantly, without capturing the temporal component of the multi-valued context, it will hard to establish frequent passing patterns of a high-passing team like FC Barcelona\protect\footnote{https://bleacherreport.com/articles/1789272-top-5-short-passing-teams-in-la-liga}, since all outfield-players may pass to one another at least once.

In this project, \textit{histogram concept-scaling} helps to ensure that the single-valued contexts capture the temporal component of the match events considered. The work by \cite{bertaux2009identifying} describes a similar procedure. Figure \ref{fig:2565917_1H} illustrates the outcome of this technique on real match data.

\begin{table}[h!]
\begin{center}
\begin{tabu} { | l | l | l | l | }
 \hline
 \textbf{\textit{Objects}} & \textbf{Messi}  & \textbf{Suarez}  & \textbf{Neymar}\\
 \hline
 Rakitic &  & $[10,15]$ & $[24,53]$\\
 \hline
 Sergio & $[4,7,60,85]$ & $[70]$ &  \\
 \hline
 Busquet & $[13,18,77]$ & $[22,34]$ &  \\
 \hline
 Pique & $[2,42]$ &  & $[8]$ \\
 \hline
 Alba & $[5,30,67]$ &  &  \\
\hline
\end{tabu}\\
\end{center}
\caption{A multi-valued formal context expressing passing relations between players.}
\label{table:MVformalcontexttoy}
\end{table}

\subsection{Implication Rules}
An attribute implication is an expression of the form \textit{X}$\rightarrow$\textit{Y}, where \textit{X},\textit{Y} $\subseteq$ \textit{M}, is true in a formal context if each \textit{m} has all attributes in \textit{X} as in \textit{Y} \cite{ganter2012formal}. It is common in relevant literature that \textit{X} is referred to as the \textit{premise} or \textit{antecedent} and \textit{Y} the \textit{conclusion} or \textit{consequent}.

Algorithms for generating implications seek to compute a subset of implications equivalent to the set of all valid implications. This subset is the \textit{cover} of valid implications.  A cover that has an \textit{infimum} is a \textit{minimum cover} of implications.  The Duquenne-Guigues (DG) basis [Guigues and Duquenne 1986] is one such cover \cite{bazhanov2014optimizations}\cite{guigues1986familles}.

The studies by \cite{stumme1999conceptual} and \cite{pasquier1998pruning} introduce a formal representation of association rules as implications, in the context of FCA. Hence, we can choose to seek interesting new bits about a set of rules (implications) and also quantify the degree to which a rule is interesting or worth considering. Popular measures of 'interestingness'\protect\footnote{https://en.wikipedia.org/wiki/Association\_rule\_learning} include \textit{support}, \textit{confidence}, and \textit{lift}. For a detailed discussions of such measures, refer to \cite{hornik2005arules} and \cite{tuffery2011data}.

\section{Methodology} \label{sec:method}

The approach commences with a scaled binary formal context built from the passes made by each FC Barcelona player at \textit{every point in time}. Hence, players (initiating a pass) constitute the set of objects, $\textit{G}$. The attribute set $\textit{M}$, is made up of pass-receiving players, whose player IDs along with the timestamp of each pass received is encoded to form attributes.
 
Secondly, the canonical base of implications, Duquenne-Guigues, is computed using an optimised Ganter Algorithm\protect\footnote{https://github.com/ae-hse/fca} \cite{bazhanov2011comparing}\cite{ganter2010two}. Like the idea of conceptual similarity expounded in \cite{aufaure2013advances}, the final step is a search for recurring patterns (i.e. similar pass sequences to an index sequence) over subsequent matches played a team. 

The recurring patterns are attribute sets, similar to or exactly as the conclusion - also an attribute set - of the index sequence.  A string similarity ratio determines what patterns make up the result and is made possible by freezing the source and target attribute sets in string data structures (stringified).

\subsection{Dataset}

The dataset\protect\footnote{Accessible online at \textit{https://figshare.com/collections/Soccer\_match\_event\_dataset/4415000/2}} is a collection that covers the top soccer leagues in Europe and two other events. Its details include match events like passes and shots, and when such events occurred \cite{pappalardo2019public}. The files from the dataset, relevant to this project are (\textit{i.}) \textit{matches\_Spain.json} (\textit{ii.}) \textit{events\_Spain.json} and (\textit{iii.}) \textit{players.json}. Of the matches played in Spain in the 2017/2018 \textit{La Liga} season by FC Barcelona, the first three were chosen to show the approach and have the following details:

\begin{itemize}
    \item Match with \textit{wyID} '2565554': FC Barcelona vs. Real Betis (2 - 0), played on August 20, 2017, at the Camp Nou.
    
    \item Match with \textit{wyID} '2565559': Deportivo Alaves vs. F.C Barcelona (0 - 2), played on August 26, 2017, at Estadio de Mendizorroza.
    
    \item Match with \textit{wyID} '2565577': FC Barcelona vs. Espanyol (5 - 0), played on September 9, 2017, at the Camp Nou.
\end{itemize}


\subsection{The Types of Passes Considered}

The types of pass events considered include those tagged as \textit{accurate}:1801, \textit{not accurate}:1802, \textit{assist}:301 and \textit{key pass}:302, after filtering the events on the team ID assigned FC Barcelona, '676'.
The opposition coaching staff would seemingly be more concerned with the lead-up to goals than the goal itself or passes in general in seeking to understand another team’s passing strategy. 
A single match event has a structure formatted in JSON as shown in the example below:

\begin{enumerate}
    \item[] \{'eventId': 8, 'subEventName': 'Simple pass', 'tags': [\{'id': 1801\}], 'playerId': 3542, 'positions': [\{'y': 61, 'x': 37\}, \{'y': 50, 'x': 50\}], 'matchId': 2565548, 'eventName': 'Pass', 'teamId': 682, 'matchPeriod': '1H', 'eventSec': 2.9945820000000083, 'subEventId': 85, 'id': 180864419\}
\end{enumerate}

A full list of the indices adopted for events and event sub-types is available at \textit{https://apidocs.wyscout.com/matches-wyid-events}.

\subsection{Scaling}

The histogram scaling scheme was implemented based on the assumption that most regular matches (other than two-legged formats like the Spanish \textit{Copa Del Rey} or UEFA competitions ) will not exceed 50 minutes in each half. Hence, consideration was only for matches fitting that assumption.

With that understanding, each relevant event’s \textit{eventSec} value - the time in seconds, into the match, when an event occurs - determines which bin of a possible 10 (indexed 0 to 9), to which an event is assigned. Each bin represents 5-minutes intervals of each half. For example, bin 0 will be assigned an event with \textit{eventSec}: 2.9945 after computing:

\begin{itemize}
     \item[] $trunc(\frac{2.9945}{60}$x$\frac{9}{50})$ = $trunc(0.0089)$ = 0 \textit{(nearest lower integer)}
\end{itemize}

\begin{figure}[ht]
\centering
\includegraphics[scale=0.40]{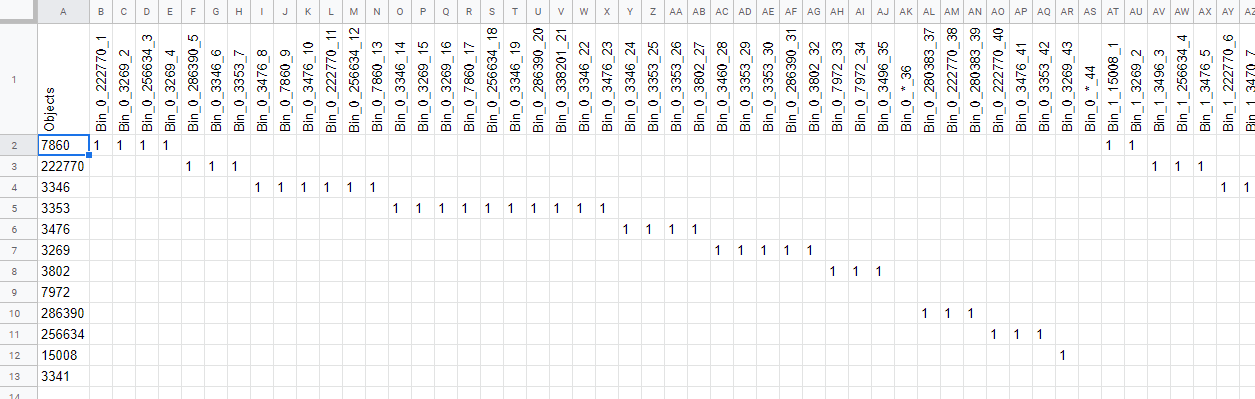}
\caption{Binary single-valued context obtained after histogram concept-scaling.}
\label{fig:2565917_1H}
\end{figure}

With this scaling scheme, a binary formal context is obtained. Figure \ref{fig:2565917_1H} shows an excerpt of the context obtained for pass events in the 1st-half of the match between Levante and F.C Barcelona (match ID '2565917'), that took place on May 13, 2018, at the Estadio Ciudad de Valencia. The context is $12$x$282$ in dimension, originally.

In Figure \ref{fig:2565917_1H}, the rows represent the players initiating passes, and the attributes (columns) encode who receives the pass and the interval of the match in which the event took place, i.e. the bin.

\subsection{Obtaining the Canonical Basis (Duquenne-Guigues)}

\begin{figure}[ht]
\centering
\includegraphics[scale=0.48]{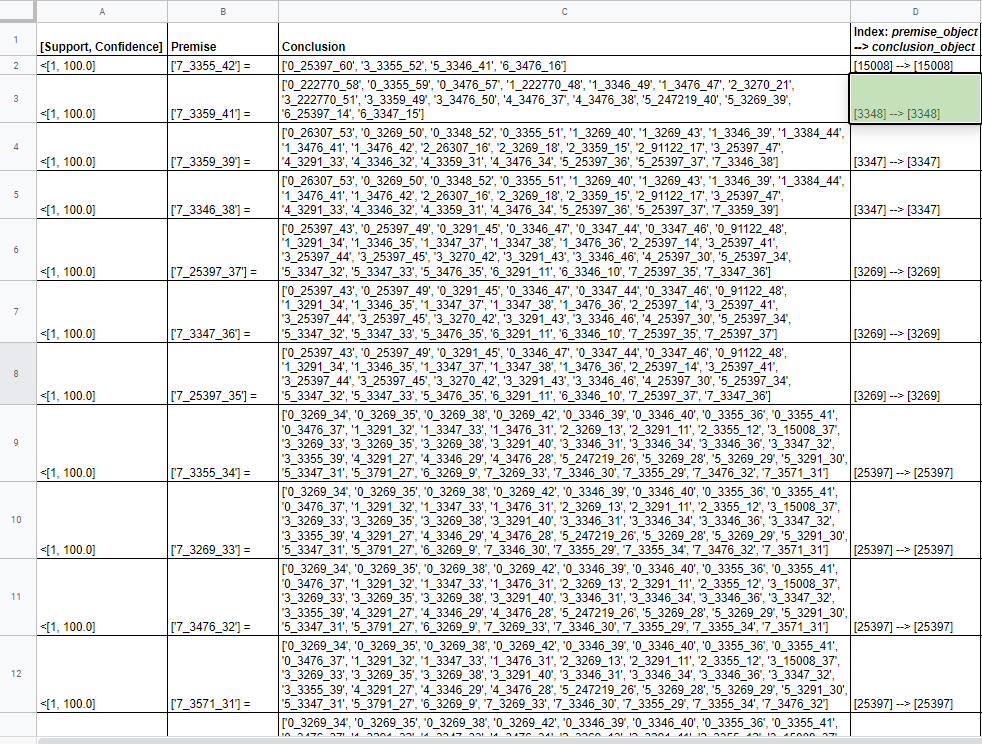}
\caption{Part of the canonical basis derived for match '2565554', 1st-half.}
\label{fig:2565554_1H}
\end{figure}

The Python code library fca (\textit{https://github.com/ae-hse/fca}) enabled the computation of the DG basis for the formal context of \textbf{each half} of the three matches considered. Figure \ref{fig:2565554_1H} shows an excerpt of the DG basis implications computed for the 2nd-half of FC Barcelona’s first match of the 2017/2018 \textit{La Liga} season. Note that the prefix 'Bin' has been muted to aid legibility.

Interestingly, the validity of the implication set holds ground as in all implications observed (after discarding those with zero support), the premise and conclusion have the same initiating player. Indicative of this is that the remaining non-zero support implications each hold a support value of 1.

\subsection{Uncovering the Strategy}

On the intuition/hypothesis that a team’s frequent passing sequences (passing itemsets) point to its passing strategy, it is only convenient to seek out any such frequent pattern in the team’s subsequent matches, with the benefit of an index match, e.g. first match of the season or a pre-season friendly.

Indexed by FC Barcelona’s first match of the 2017/2018 season, specifically the 1st-half play, the conclusions of the DG basis implications generated for that half were used in a search for similar occurrences, firstly, against the implication conclusions of the 2nd-half of the index match itself, and then against the implication conclusions of each half of the team’s following two matches. The two matches being those played in game-weeks 2 (match '2565559') and 3 (match '2565577').

The Python list structures representing the index match (’ 2565554’) conclusions were 'stringified' - wrap a Python \textit{str()} function around a \textit{list(list\_object)}  -  and a string similarity ratio (e.g. Levenshtein distance) then computed against each conclusion (also stringified) in the target matches.

Using the  Python package \textit{fuzzy-wuzzy} (\textit{https://github.com/seatgeek/fuzzywuzzy}) to measure string similarity suited the use-case, given its \textit{process.extract("query", $[choice_{1},...,choice_{n}]$)} function. Specifically, a single setting was enforced in all cases: \textit{process.extract("query", $[conclusion_{1},...,conclusion_{n}]$, scorer=fuzz.token\_sort\_ratio, score\_cutoff=75, limit=10)}. The limit parameter dictates the maximum number of results returned. The \textit{score\_cutoff} parameter limits the result to those patterns with score/ratio less than the assigned value.

In Figure \ref{fig:2565554_1H}, the implication conclusions reflect the fact that a player can pass to the same player more than once and within the same time frame (bin), i.e. a receiving player’s ID can appear in the target implication conclusion more than once. Therefore, in computing the similarities, that possibility is addressed by parsing each target conclusion (stored as Python lists) through a Python set and then sorting: \textit{sorted(set(conclusion))}. Doing this keeps the search in the proper perspective, i.e.\textbf{we are interested in those patterns in which a crop of players, possibly exactly as those glimpsed in the query conclusion, exist in the target conclusion}.

\section{Results} \label{sec:results}

\begin{figure}[ht]
\centering
\includegraphics[scale=0.50]{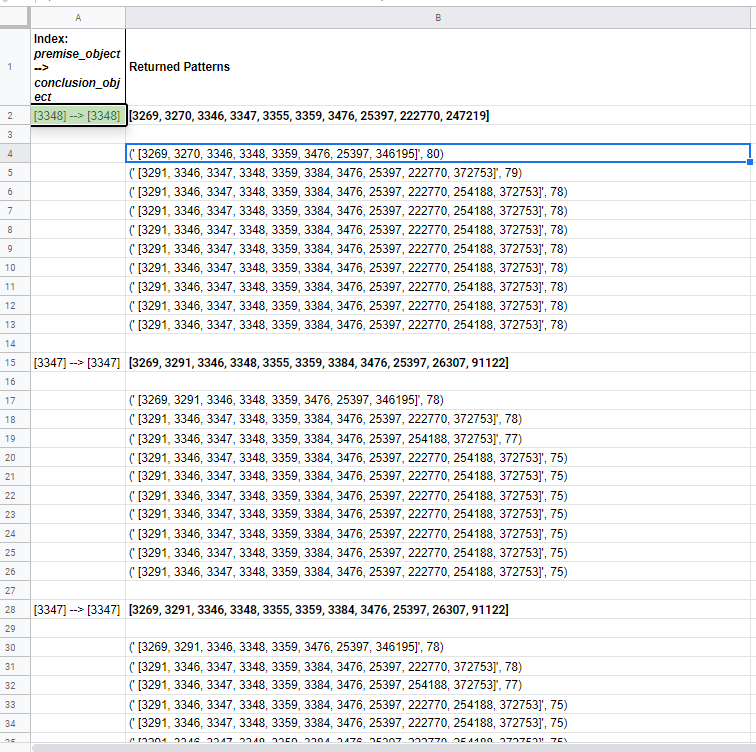}
\caption{Search results for index match 2565554\_1st-half against \textit{self}\_2nd-half. The cell highlighted in green in \textit{Fig. \ref{fig:2565554_1H}} is now shown here, next to a list of passing sequences similar to the conclusion it indexes and their similarity ratios.}
\label{fig:2565554_1H_similars_SELF_2H}
\end{figure}

An excerpt of the patterns (similar passing sequences) has been presented as the result of the approach, for brevity. See Figure \ref{fig:2565554_1H_similars_SELF_2H}. The result shows that it is possible to deduce the passing strategy of a soccer team, by studying the team’s passing sequences, formalised as implication rules, over several matches.

Some post-processing step can be applied to improve the duplicity in the query results. The total number of  DG basis implications with non-zero support generated from the three matches analysed was 1,915. Undoubtedly, much more can be done in future research efforts. For further enquiry, the full set of results is publicly available\protect\footnote{https://docs.google.com/spreadsheets/d/1e5WBQkpK\_uIPs5wBtWw4r03Yrz8h2dPhwusYYOJ-18g/edit?usp=sharing}.

\section{Conclusion}

This project has demonstrated a formal concept analysis-based approach to uncovering the passing strategy of a soccer team, using passing events in soccer play-by-play data made openly available by \cite{pappalardo2019public}. The importance of open-access data cannot be overemphasised. 

The implication set described in this project fully supports further data-mining tasks, for example, knowledge exploration by employing Armstrong’s rules of inference.  Hopefully, other researchers can take up such an effort and more.
\newpage

\bibliographystyle{sbc}
\bibliography{sbc-template}

\end{document}